\documentclass[letterpaper, 10 pt, conference]{ieeeconf}  % Comment this line out if you need a4paper
\usepackage{amssymb}
\usepackage{amsmath}
\usepackage[margin=1in]{geometry}
\usepackage{booktabs}  % 若想要更美观的表格线，可以使用 booktabs 宏包

\IEEEoverridecommandlockouts                              % This command is only needed if 
\title{\LARGE \bf
ExoGait-MS: Learning Periodic Dynamics with Multi-Scale Graph Network for Exoskeleton Gait Recognition}

\author{Lijiang Liu$^{1}$ Junyu Shi$^{1}$, Yong Sun$^{1}$, Zhiyuan Zhang$^{1}$,  Jinni Zhou$^{1}$, Shugen Ma$^{1}$, Qiang Nie$^{1,*}$% <-this % stops a space
\thanks{$^{1}$ROAS Thrust, System Hub,  Hong Kong University of Science and
 Technology (Guangzhou), Guangdong, China.
        }%
\thanks{$^{*}$Corresponding author.
        {\tt\small qiangnie@hkust-gz.edu.cn}}%
}

\begin{document}

\maketitle
\thispagestyle{empty}
\pagestyle{empty}

%%%%%%%%%%%%%%%%%%%%%%%%%%%%%%%%%%%%%%%%%%%%%%%%%%%%%%%%%%%%%%%%%%%%%%%%%%%%%%%%
\begin{abstract}

Current exoskeleton control methods often face challenges in delivering personalized treatment. Standardized walking gaits can lead to patient discomfort or even injury.  Therefore, personalized gait is essential for the effectiveness of exoskeleton robots, as it directly impacts their adaptability, comfort, and rehabilitation outcomes for individual users. To enable personalized treatment in exoskeleton-assisted therapy and related applications, accurate recognition of personal gait is crucial for implementing tailored gait control. The key challenge in gait recognition lies in effectively capturing individual differences in subtle gait features caused by joint synergy, such as step frequency and step length. To tackle this issue, we propose a novel approach,  which uses Multi-Scale Global Dense Graph Convolutional Networks (GCN) in the spatial domain to identify latent joint synergy patterns. Moreover, we propose a Gait Non-linear Periodic Dynamics Learning module to effectively capture the periodic characteristics of gait in the temporal domain. To support our individual gait recognition task, we have constructed a comprehensive gait dataset that ensures both completeness and reliability. Our experimental results demonstrate that our method achieves an impressive accuracy of 94.34\% on this dataset, surpassing the current state-of-the-art (SOTA) by 3.77\%. This advancement underscores the potential of our approach to enhance personalized gait control in exoskeleton-assisted therapy.

\end{abstract}

%%%%%%%%%%%%%%%%%%%%%%%%%%%%%%%%%%%%%%%%%%%%%%%%%%%%%%%%%%%%%%%%%%%%%%%%%%%%%%%%
\section{INTRODUCTION}

Exoskeleton robotic technology\cite{nazari2023applied}\cite{doi:10.1177/16878140211011862} has significantly advanced rehabilitation\cite{cullen2012reaffirming} by enhancing mobility recovery in individuals with gait impairments. However, most existing exoskeleton systems rely on standardized gait patterns, which may not adequately accommodate the diverse and individualized locomotor characteristics of patients. Over time, individuals develop unique gait patterns influenced by their musculoskeletal adaptations, accumulated fatigue, and underlying pathological conditions. Consequently, applying a uniform gait strategy across all users may not only reduce rehabilitation efficacy but also pose risks of discomfort or secondary injury. To optimize rehabilitation outcomes, it is imperative to develop exoskeleton control strategies that offer personalized gait assistance. This necessitates an accurate identification of an individual’s gait characteristics, enabling adaptive control mechanisms that can replicate or complement their natural movement patterns. Therefore, one of the fundamental challenges in this domain is robust gait recognition—ensuring that the exoskeleton system can reliably distinguish, analyze, and respond to individual gait patterns. This study aims to address this challenge by focusing on gait recognition as the foundational step toward personalized exoskeleton-assisted rehabilitation. By establishing a reliable framework for identifying and classifying individualized gait patterns, this work lays the groundwork for the development of adaptive control strategies, ultimately enhancing user-specific rehabilitation and promoting long-term mobility improvements.
%上面这一段是为了替换introduction的 这是备选第二段 明早来了看看和第一个备选 哪个更合适

%Currently, exoskeleton robots face limitations in terms of personalization, with gait generation %primarily relying on individualized patient data or predefined gait parameters. For instance, the %trajectory generation of Lokomat requires adjustments based on the patient's stride length and %individual characteristics. At the same time, the LOPES method\cite{7548389} reconstructs gait %trajectories using regression analysis based on height and speed. These %methods\cite{luo2022trajectory} exhibit a strong dependence on data. Although statistical learning %methods\cite{vapnik1999overview}, such as radial basis function neural %networks\cite{lee1999robust}and multilayer perceptron neural networks\cite{desai2021anatomization}, %can be used for motion prediction when dealing with large datasets, they rely on a substantial %amount of samples and lack the flexibility for real-time adjustments. Furthermore, gait planning %methods that depend on predefined gait parameters, such as model predictive control %%%(MPC)\cite{garcia1989model} for online trajectory generation, while capable of producing gait %patterns based on predefined parameters, still lack personalization and dynamic adaptability. These %methods fail to simulate individual differences fully and are insufficient in providing real-time %feedback in complex environments.

Personalized gait recognition presents significant challenges in the context of exoskeleton-assisted control. While gait patterns exhibit general similarities across individuals, subtle variations exist at the micro-level, including stride length, step frequency, joint trajectories, and coordination dynamics. The complexity of these individualized characteristics necessitates not only the modeling of temporal dynamics but also the precise capture of latent inter-joint coordination and their sequential dependencies throughout the gait cycle. Moreover, gait is a highly dynamic process influenced by biomechanics, neuromuscular control, and external environmental factors, exhibiting strong nonlinearity and inter-individual variability. These factors make traditional approaches, which rely on static features or predefined rules, inadequate for personalized gait modeling and adaptation. Existing research in gait recognition has made significant progress, primarily focusing on extracting gait features from RGB videos, depth information, or inertial sensor data and applying statistical learning or deep learning techniques for classification and recognition. 

However, these methods predominantly aim at recognizing generic gait patterns rather than achieving precise individualized gait modeling for exoskeleton control optimization. Furthermore, current mainstream approaches often overlook the dynamic coordination characteristics of gait, failing to effectively model the implicit inter-joint interactions throughout gait evolution. This limitation hinders their applicability in real-time personalized exoskeleton assistance. Therefore, there is a pressing need for a method capable of accurately extracting personalized gait features while establishing a direct mapping between gait patterns and exoskeleton control strategies. Such an approach would enhance the adaptability and generalization of exoskeleton systems in real-time gait assistance tasks.

%In this paper, we propose a vision-only gait recognition framework that operates on RGB-captured walking sequences, eliminating reliance on specialized sensors or labeled datasets. Our method begins by extracting skeleton sequences directly from raw video streams, which intuitively preserves individual gait characteristics while reducing domain-specific data dependency. To address the challenge of inter-individual variability (e.g., differences in step length, cadence, and motion scales), we introduce a dual-branch architecture: one branch learns nonlinear periodic dynamics using an encoder-decoder structure in low-dimensional space, while the other employs cross-scale global graph convolutions to flexibly model global joint dependencies and spatial structural signatures. By integrating these complementary spatiotemporal representations, our framework achieves promising gait recognition accuracy.

% In this paper, we propose a vision-based gait recognition method, which collects human walking data through RGB cameras and extracts skeleton sequences. This approach reduces reliance on specific data and more intuitively captures individual gait characteristics. Through this method, we can more accurately extract personalized gait trajectories without excessive dependence on external gait parameters, thus providing more flexible and individualized gait services. This approach offers strong real-time performance and dynamic adaptability, making it an effective complement to current gait generation techniques. 

In this paper, we present a comprehensive framework for gait recognition that simultaneously captures both temporal dynamics and spatial features of human gait. As depicted in Figure \ref{figure_pipeline}, our approach is structured with two parallel branches. The first branch focuses on learning the temporal gait dynamics by utilizing an encoder-decoder architecture, which effectively captures the nonlinear periodic patterns of gait in a reduced latent space. The second branch emphasizes spatial feature extraction through multi-scale global graph convolutional layers (MGD-GCL), which model the dependencies among joints at multiple scales to understand the motion patterns across the human body. These temporal and spatial features are fused in a spatiotemporal fusion layer, which integrates both dimensions to form a comprehensive representation of gait. As shown in Figure \ref{fig:your_label}, the process of extracting 3D gait information starts with 2D pose estimation from raw video input, followed by a 2D-to-3D lifting model to reconstruct the 3D skeletal structure. This approach reduces the reliance on manual labeling by automating the 3D pose reconstruction, thereby enhancing the scalability and efficiency of the gait analysis process.xperimental results demonstrate that our method achieves 94.34\% accuracy on this dataset, outperforming the state-of-the-art (SOTA) by 3.77\%.

\section{RELATED WORKS}

\subsection{Graph Convolutional Networks for Sequential Skeleton Data}
Graph Convolutional Networks (GCNs)\cite{kipf2017semisupervised} have become a cornerstone in skeleton-based human motion-related tasks, such as action recognition\cite{duan2022revisiting,liu2020disentangling}, motion prediction\cite{shi2024multi,xu2023eqmotion,li2021multiscale} and pose estimation\cite{zhai2023hopfir,wu2021graph}, capitalizing on their ability to model the spatial relationship of unstructured body joints. However, the original GCNs only focus on spatial information transmission, ignoring the temporal features. To break this limitation, STGCN\cite{yu2018spatio} proposes spatio-temporal GCNs to simultaneously capture the spatiotemporal characteristics of motion sequences, thereby addressing both the connectivity of joints and their evolution over time. LTD\cite{mao2019learning} transforms the sequence skeleton data from the temporal domain to the DCT domain, implicitly modeling temporal features. In addition, temporal structure can also be viewed as a time graph by connecting adjacent frames. Based on this, MSTGNN\cite{li2021multiscale} extracts spatiotemporal features by using GCNs alternately for space graphs and time graphs. In summary, these advances innovatively address the integration of temporal dynamics in skeleton-based motion analysis. 

\subsection{Gait Recognition}
Gait Recognition, as a key task in biometrics and human motion analysis, has gained significant attention. In exoskeletons, for example, tailored assistance and optimized rehabilitation outcomes are provided by accurately recognizing individual gait patterns. Gait recognition methods can generally be classified into appearance-based\cite{tanawongsuwan2004modelling} and model-based\cite{zhang2020learning,fan2024skeletongait,pinyoanuntapong2023gaitmixer} approaches. Appearance-based methods focus on body shape from a visual perspective, using silhouettes or gait analysis sequences as inputs. In contrast, model-based approaches emphasize the intrinsic structure of the human body, initially utilizing bones and later evolving to incorporate meshes. Recently, gait recognition based on image modality\cite{zhang2020learning}, skeleton modality\cite{teepe2021gaitgraph,pinyoanuntapong2023gaitmixer,fan2024skeletongait}, and 3D point cloud modality\cite{shen2023lidargait} has become mainstream for model-based methods. However, the image modality is easily affected by the clothing of the subject, 3D point cloud data is relatively difficult to obtain, and appearance data lacks fine granularity. In contrast, gait recognition methods based on skeleton modality have the advantages of being less affected by interfering factors and having high precision in capturing movements.

This paper focuses on the skeleton-based gait recognition. GaitGraph\cite{teepe2021gaitgraph} applies GCN to extract gait features, incorporating powerful spatiotemporal modeling. Based on GaitGraph, GaitGraphv2\cite{teepe2022towards} combines higher-order inputs, and residual networks to an efficient method for gait recognition. GaitMixer\cite{pinyoanuntapong2023gaitmixer} follows a heterogeneous multi-axis mixer architecture by first employing a spatial self-attention mixer and then a large-kernel temporal convolution mixer, thereby extracting rich multi-frequency signals from the gait feature maps. Rather than directly model the skeleton graph, SkeletonGait\cite{fan2024skeletongait} represents the coordinates of human joints as a heatmap with Gaussian approximation, exhibiting a silhouette-like image devoid of exact body structure. However,these methods ignore the nonlinear periodic features of human moving patterns and the cross-scale relationships of global human body joints, which are the key points for analyzing the gait features in this paper.

\section{Method}

\subsection{Dataset Construction}
\begin{figure}[t]
\includegraphics[width=1.0\linewidth]{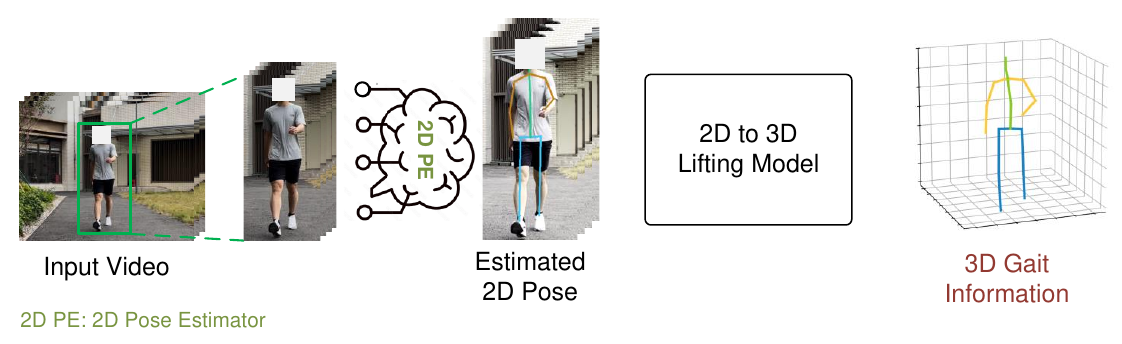} 
 \caption{Pipeline for extracting 3D gait information from an input video using a 2D pose estimator and a 2D-to-3D lifting model.}
     \label{fig:your_label}
\end{figure}

\begin{figure}[h]
    \centering
    \includegraphics[width=0.45\textwidth]{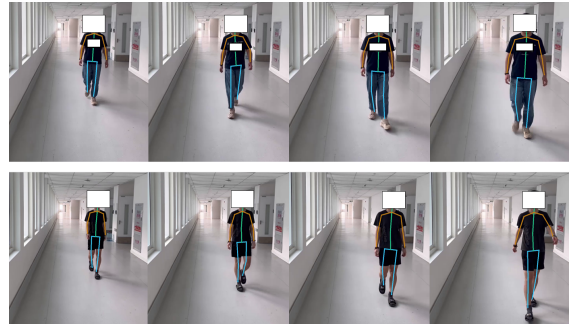} % 修改为你的图片
    \caption{This is an example of detection results with 3D keypoints for different individuals.} % 图片标题
    \label{fig:3d} % 用于引用的标签
\end{figure}

This study employs a self-collected gait dataset, which includes gait data from 24 participants. Each participant undergoes gait recording in five distinct environments, with eight gait sequences captured per environment. During data collection, participants walk according to their habitual gait patterns, with personalized gait adjustments made to account for individual characteristics. The recorded video data undergoes a systematic preprocessing pipeline, involving the extraction of 2D keypoints from raw video frames and the subsequent reconstruction of 3D skeletal representations through a 2D-to-3D lifting technique. This automated annotation process significantly reduces the reliance on manual labeling, thereby enhancing both efficiency and scalability. Ultimately, a structured skeletal sequence dataset is obtained to support subsequent modeling and analysis. The pipeline for obtaining 3D skeletal data from video is shown in Fig. \ref{fig:your_label}. The results of extracting skeletal data from different individuals are shown in Fig.  \ref{fig:3d}.

%\begin{figure}[h]
%    \centering
 %   \includegraphics[width=0.45\textwidth]{gait2.png} % 修改为你的图片
%    \caption{Illustration of the proposed method.} % 图片标题
%    \label{fig:method} % 用于引用的标签
%\end{figure}

\begin{figure*}[t!]
\includegraphics[width=1.0\linewidth]{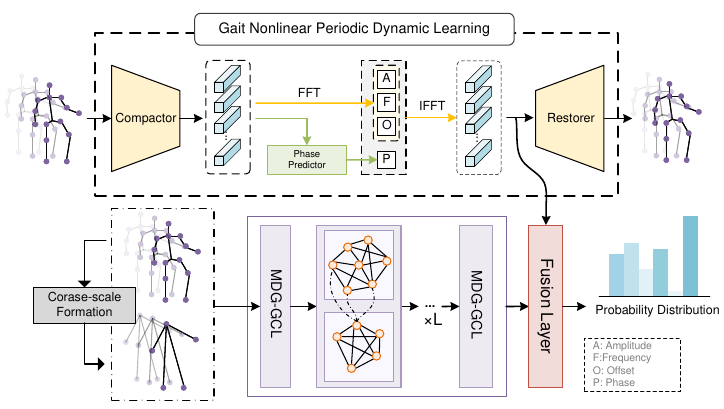} 
\caption{The architecture of our proposed gait recognition framework consisting of two parallel branches. The top branch learns temporal gait dynamics via an encoder-decoder that captures nonlinear periodic patterns in a low-dimensional latent space, while the bottom branch extracts spatial gait features through cross-scale global graph learning, explicitly modeling joint-wise dependencies.}
    \label{figure_pipeline}
\end{figure*}

\subsection{Gait Recognition}
As shown in Fig. \ref{figure_pipeline}, our method contains the concurrent modeling of gait features in both temporal and spatial dimensions. Drawing from inspiration from periodic autoencoders and frequency latent dynamics\cite{starke2022deepphase,starke2023motion}, we incorporate a differentiable Fast Fourier Transform (FFT) and a phase learning mechanism within the temporal domain to effectively parameterize the intrinsic periodic characteristics of gait motion sequences to learn the gait dynamics. In the spatial domain, we utilize multi-scale dense global graph convolutional layers (MGD-GCL) to concurrently capture adaptable latent relationships among joints and discern motion patterns from human body graph representations at varying scales. Finally, through a spatiotemporal fusion layer, we embed the periodic representation of gait into the latent vector of joints.

\subsubsection{Gait Non-linear Periodic Dynamics Learning}
Human gait data, though inherently a high-dimensional time series, is shaped by the physical constraints of joint movements, resulting in a low-dimensional manifold structure. Inspired by this insight, we employ an autoencoder to explicitly extract low-dimensional periodic features from unstructured gait data.

Given a gait sequence \( \mathbf{X} = \left[ x_1, x_2, ..., x_T \right] \) with \( x_t \in \mathbb{R}^d \), where \( T \) is the number of frames and \( d \) is the total number of joint trajectories along the \( x,y,z \) axis. To capture gait dynamics, we compute the velocity sequence by diferencing consecutive frames:
\begin{equation}
    \mathbf{V}_{t}=\mathbf{X}_{t+1}-\mathbf{X}_{t}, \quad t=1,2, \ldots, T-1
\end{equation}
To capture the nonlinear coupling in joint movements—such as the relationship between upper limb swing and stride frequency—we first utilize a trainable compactor to project the motion data into a low-dimensional latent representation:
\begin{equation}
    \mathbf{H} = \mathcal{G}(\mathbf{V})
\end{equation}
where \( \mathcal{G} \) is consist of two convolutional layers with activation function. Then, we parameterize each latent curve derived from \( \mathbf{H} \) by applying the differentiable real FFT to each channel of it along the time dimension to obtain amplitude, frequency, and offset. The FFT produces a matrix of Fourier coefficients \( \mathbf{c} \in \mathbb{R}^{K \times (\left\lfloor\frac{T}{2}\right\rfloor + 1)} \), defined as \( \mathbf{c}_{i} = \operatorname{FFT}(\mathbf{H}_{i,:}) \). 
% The per-channel power spectrum is then computed as:
% \begin{equation}
%  \mathbf{p}_{i, j}=\frac{\mathbf{2}}{T}\left|\mathbf{c}_{i, j}\right|^{2}, \quad j = 1, 2, ..., \left\lfloor\frac{T}{2}\right\rfloor
% \end{equation}
Then, the amplitude \( \mathbf{A} \), frequency \( \mathbf{F} \) and offset \( \mathbf{O} \) can be expressed as:
\begin{equation}
    \boldsymbol{A}_{i}=\sqrt{\frac{4}{T^{2}} \sum_{j=1}^{\left\lfloor\frac{T}{2}\right\rfloor}\left|\mathbf{c}_{i, j}\right|^{2}}
\end{equation}
\begin{equation}
    \mathbf{F}_{i}=\frac{\sum_{j=1}^{\left\lfloor\frac{T}{2}\right\rfloor} f_{j} \cdot\left|\mathbf{c}_{i, j}\right|^{2}}{\sum_{j=1}^{\left\lfloor\frac{T}{2}\right\rfloor}\left|\mathbf{c}_{i, j}\right|^{2}}, \quad f_{j}=\frac{j}{N} 
\end{equation}
\begin{equation}
    \mathbf{O}_{i}=\frac{\operatorname{Re}\left(\mathbf{c}_{i, 0}\right)}{T}
\end{equation}
where \( N \) represents the length of the window slide, \( \operatorname{Re} \) stands for the real part of the \( \mathbf{c}_i \).

To obtain the phase shifts, for each channel, the input \( \mathbf{H}_i \) is first processed through a phase predictor \( \mathcal{P} \) to produce a two-dimensional vector \( q_i = {q_{i, 0}, q_{i, 1}} \). The phase angle \( \theta_i \) is subsequently calculated by:
\begin{equation}
    \theta_i = \operatorname{arctan2}(q_i), \quad \theta_i \in [-\pi, \pi]
\end{equation}
After normalization \( \mathbf{p}_i = \frac{\theta_i}{2\pi} \), we obtain the phase.
To train the autoencoder, we reconstruct the latent gait signal by:
\begin{equation}
    \hat{\mathbf{H}} = \mathbf{A}\operatorname{sin}(2\pi(\mathbf{F}N + \mathbf{P})) + \mathbf{O}
\end{equation}
Finally, we use the restorer to decode the latent signal, resulting in the frame-level gait sequence \( \mathbf{X}_p \).

\subsubsection{Multi-Scale Global Dense Graph Convolutional Layer}
We propose the multi-scale globel dense graph convolutional layer to capture both detailed joint-level features and higher-level abstract features, using a combination of fine-scale and coarse-scale graphs, and introduces a trainable adjacency matrix to enhance flexibility and global connectivity. The fine-scale graph is defined as the original human body graph, where the coarse-scale graph simplifies the representation by averaging adjacent nodes. For instance, we average the right wrist joint, right elbow joint and right shoulder joint to obtain the right leg coarse representation.

We apply GCN to the human graph modeling. To capture the flexible relationships between joints and the cross-scale relationships between joints and body part, we treat the human graph as a dense graph, that each joint is connected with all the others (including cross-scale connections). We use a trainble, randomly initialized adjacency matrix \( \mathbf{A}_{train} \in \mathcal{R}^{(R_1 + R_2) \times (R_1 + R_2)} \) instead of the Laplacian-normalized adjacency matrix based on the graph's fixed topology, where \( R_1 \) and \( R_2 \) are the number of joints in fine-scale and coarse-scale human graph structure. This process in the layer \( l \) can be expressed as follows:
\begin{equation}
    \mathbf{h}_{i}^{(l+1)}=\sigma\left(\sum_{j=1}^{R_1 + R_2} \mathbf{A}_{train, i j}^{(l)} \mathbf{W}^{(l)} \mathbf{h}_{j}^{(l)}\right)
\end{equation}
where \( \sigma \) is the activation function, \( \mathbf{W}^{(l)} \) is the trainable weight matrix at layer \( l \), \( \mathbf{h}_{i}^{(l+1)} \) denotes the feature vector of joint \( i \) at layer \( l+1 \).

\subsubsection{Spatiotemporal Fusion Layer}
The reconstructed latent signal derived from the phase manifold is fused with the output features from the MGD-GCLs, which creates a combined feature vector that retains both temporal and spatial information. Then, we apply a single MGD-GCL to produce the output probability distribution \( \mathbf{p} \).

\subsubsection{Loss Function}
We use Mean Squared Error (MSE) to measure the difference between the phase manifold's reconstructed signal \( \mathbf{X}_m \) and the original signal \( \mathbf{X} \):
\begin{equation}
    \mathcal{L}_{MSE} = \frac{1}{T}\sum_{i=1}^{T}\left\|\mathbf{X}^i-\mathbf{X}_{m}^i\right\|_{2}^{2}
\end{equation}
For the \( z\)-class gait recognition training, we use cross-entropy loss to optimize the probability distribution \( \mathbf{p} \):
\begin{equation}
    \mathcal{L}_{CE} = -\sum_{i=1}^{z} y_{i} \log \left(\mathbf{p}_{i}\right)
\end{equation}

%\begin{figure}[ht]
%    \raggedright
 %   \includegraphics[width=1.0\linewidth]{绘图1.png}
  %  \caption{The overall flowchart of the idea}
   % \label{fig:your_label}
%\end{figure}
%\begin{figure}[ht]
%\includegraphics[width=1.0\linewidth]{绘图2.png} 
% \caption{Detailed local logic}
%     \label{fig:your_label}
%\end{figure}

\section{Experiments}

\subsection{Implementation Details}

 The 2D pose estimation module adopts YOLOv8-Pose architecture, with 3D pose reconstruction achieved through a 12-layer Hotplate-Transformer (HoT). The GCN backbone utilizes 512-dimensional hidden representations. Training is conducted on an NVIDIA RTX 3060 GPU with PyTorch 2.1.0, using Adam optimizer under cosine-annealed learning rate scheduling (initial lr=5e-4, batch size=16) over 100 epochs. This configuration balances computational efficiency with model performance while ensuring training stability.

\subsection{Evaluation Metric}

To evaluate the performance of our approach, we employ three key classification metrics: 
1. AUC (Area Under Curve) measures the model’s ability to distinguish between classes.
2. Accuracy calculates the proportion of correctly classified samples.
3. F1-Score balances precision and recall for imbalanced data evaluation.
%太笼统了，这里要具体去写 每个评教指标的作用。详细写 不笼统

1. Area Under Curve (AUC)
AUC measures the model's ability to distinguish between positive and negative samples. It is computed from the Receiver Operating Characteristic (ROC) curve, where:
\begin{align}
    AUC &= \int_{0}^{1} TPR(FPR) \, d(FPR)
    %\label{eq:AUC} \\
    %&= \sum_{i=1}^{N} P_i \label{eq:sum}
\end{align}

True Positive Rate (TPR) / Recall:$$\frac{T P}{T P+F N}$$

False Positive Rate (FPR):$$\frac{F P}{F P+T N}$$
A higher AUC indicates better discriminative power.

2. Accuracy
Accuracy measures the proportion of correctly classified samples:\begin{equation}
\text{Accuracy} = \frac{TP + TN}{TP + TN + FP + FN}
\end{equation}
%where 

%\hspace{-2ex}TP (True Positives): Correctly predicted positive samples \\
%TN (True Negatives): Correctly predicted negative samples \\
%FP (False Positives): Incorrectly predicted positive samples \\
%FN (False Negatives): Incorrectly predicted negative samples

  3. \textbf{F1-Score} F1-Score balances precision and recall, particularly useful in imbalanced datasets.

$$
F 1=2 \times \frac{\text { Precision } \times \text { Recall }}{\text { Precision }+ \text { Recall }}
$$
where
\begin{align}
    &\text{} \notag \\
    &\text{Precision} = \frac{TP}{TP + FP} \\
    &\text{Recall} = \frac{TP}{TP + FN}
\end{align}
%Comparison

\subsection{Comparison}
\paragraph{Baseline}  \par
To verify the effectiveness of our gait recognition approach, we compared it with state-of-the-art (SOTA) methods, including ST-GCN\cite{yu2018spatio}, MSST-GCN\cite{chen2021multi}, GaitMixer\cite{pinyoanuntapong2023gaitmixer}, and De-GCN\cite{myung2024degcn}. Among these, ST-GCN extends graph neural networks to a spatiotemporal graph model, simultaneously extracting the spatial and temporal dependencies of actions. MS-GC replaces traditional spatial graph convolutions with subgraph convolutions, using a hierarchical structure and residual connections to capture both short- and long-range joint relationships. Meanwhile, MT-GC extends this approach to the temporal domain to model dependencies across distant frames. Their integration forms the multi-scale spatiotemporal graph convolutional network (MST-GCN). GaitMixer employs a heterogeneous multi-axis mixer architecture, utilizing a spatial self-attention mixer and a temporal large-kernel convolutional mixer to learn the rich multi-frequency signals in gait feature maps. DeGCN adaptively captures the most informative joints by learning deformable sampling locations on both spatial and temporal graphs to perceive discriminative receptive fields. At the same time, it employs a multi-branch framework to balance accuracy and model size and to enhance the integration of joint and bone modalities.
%介绍别人的方法怎么做的：为了验证我们xxxx有效性，我们和SOTA方法进行了比较，包括xxxxxxx、xx、xx。其中STGCN用了xxx方法；MSST-GCN是xxxxx；GaitMixerxxxx。

\paragraph{Quantization Comparison and Analysis} % 我们在表1中展示了xxxxx。其中我们的方法在xxxxx指标上达到了最好，分别高于SOTA方法xxx%。Acc是xxxx， F1-score。因为我们更加关注非线性周期动态特征，这一特征在xxxx，所以我们的方法有更好的xxxxx。相比于DeGCN，我们使用XXXX的GCN，关注了xxxx的方面的内在联系，所以xxxx。
%我们在我们建立的数据集上，把我们的方法和别的方法进行对比。结果如表1所示.我们在表1中展示了xxxxx。其中我们的方法在xxxxx%指标上达到了最好，分别高于SOTA方法xxx%。Acc是xxxx， F1-score是xxx。因为我们更加关注非线性周期动态特征，这一特%征在步态上有很明显的xxxx性，所以我们的方法有更好的xxxxx。相比于之前的方法，我们提出的的MDG-GCN，关注了跨尺度全局关节关系的交互，所以xxxx。
We compare our proposed method with existing approaches on our constructed dataset. The results are shown in Table 1. Our method achieves the best performance on all evaluation metrics, outperforming the state-of-the-art (SOTA) method significantly. Specifically, the accuracy (Acc) is 94.34\%, and the F1-score is 0.9428. We attribute this to our focus on capturing non-linear periodic dynamic features, which exhibit significant periodic dynamic characteristics in gait. Compared to previous approaches, the proposed MDG-GCN explicitly models cross-scale global joint interactions, leading to enhanced performances.

\begin{table}[t!] % t! 强制表格靠顶部
    \centering
    \renewcommand{\arraystretch}{1.2}
    \setlength{\tabcolsep}{8pt}
    \caption{Comparison of gait recognition models on our dataset. Our model outperforms others in terms of Accuracy, F1-score, and AUC}
    \label{tab:performance}
    \begin{tabular}{l|ccc}
        \toprule
        \textbf{Model} & \textbf{Accuracy (\%)} & \textbf{F1-score} & \textbf{AUC} \\
        \hline
        MLP-based   & 44.03  & 0.3982  & 0.9020 \\
        STGCN       & 86.16  & 0.8586  & 0.9970 \\
        MSST-GCN    & 64.15  & 0.6195  & 0.9589 \\
        GaitMixer   & 81.76  & 0.8035  & 0.9827 \\
        DeGCN       & 90.57  & 0.8988  & 0.9981 \\
        \hline
        Ours        & \textbf{94.34}  & \textbf{0.9428}  & \textbf{0.9990} \\
        \bottomrule
    \end{tabular}
\end{table}

\paragraph{Visualization of Learned joint relations}

\begin{figure}[t]
    \centering
    \includegraphics[width=0.48\textwidth]{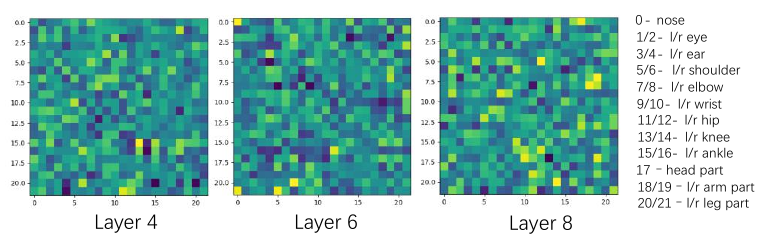}
    \caption{Visualization of the learned adjacency matrix across different graph layers.}
    \label{figure_vis} % 用于引用的标签
\end{figure}

We visualized the adjacency matrices obtained after training different graph convolutional layers (specifically the 4th, 6th, and 8th layers) in Fig. \ref{figure_vis}. In these visualizations, brighter values indicate stronger correlations between the corresponding node pairs.

From the results of the 4th layer, we observe that in the context of gait recognition, the motion patterns of the left and right knees exhibit a strong correlation, characterized by similar speeds but opposite directions. Additionally, there is a notable association between the left hip (fine scale) and the left leg part (coarse scale), representing hidden features extracted by the model.

Turning to the 8th layer, we identify a significant cross-scale correlation between the right knee (fine scale) and the left leg part (coarse scale). Furthermore, the right hip and right ankle jointly capture the internal motion pattern of the right leg. A broader trend emerges across the layers: in the shallower layers, the model primarily focuses on the motion of specific joint pairs. As the model deepens, it increasingly emphasizes cross-scale relationships and interactions between joints from different body parts. This evolution underscores the effectiveness of our proposed MDG-GCN in capturing complex dependencies for gait recognition.

\subsection{Ablation Study}
We conducted ablation experiments on the number of MDG-GCN layers and the non-linear periodic dynamic learning to validate the effectiveness of our key designs.

\begin{table}[t]
\centering
\label{table_1}
\renewcommand{\arraystretch}{1.2}
\setlength{\tabcolsep}{8pt}
\caption{Performance comparison with and without temporal information on our dataset. The model with Temp, achieves better results}
\label{table_temp}
\begin{tabular}{c|ccc}
\toprule
    & Accuracy(\%)      & \multicolumn{1}{c}{F1-score} & \multicolumn{1}{c}{AUC} \\ \hline
w/o Temp. & 89.94   & 0.8953                        & 0.9978                  \\
w/ Temp.  & \textbf{94.34} & \textbf{0.9428}        & \textbf{0.9990}                 \\ 
\bottomrule
\end{tabular}
\end{table}

\begin{table}[t]
\centering
\renewcommand{\arraystretch}{1.2}
\setlength{\tabcolsep}{8pt}
\caption{Performance comparison of GCN with different layers on the gait recognition task. The 12-layer GCN achieves the best performance}
\label{table_gcn}
\begin{tabular}{c|ccc} 
\toprule
Layers & Accuracy (\%) & F1-score & AUC \\
\hline
2  & 81.13 & 0.8082 & 0.9623 \\
4  & 85.53 & 0.8449 & 0.9597 \\
6  & 89.31 & 0.8900 & 0.9753 \\
8  & 90.57 & 0.8992 & 0.9801 \\
10 & 92.45 & 0.9203 & 0.9729 \\
12 & \textbf{94.34} & \textbf{0.9428} & \textbf{0.9990} \\
14 & 91.19 & 0.9051 & 0.9795 \\
\bottomrule
\end{tabular}
\end{table}

\paragraph{Number of MDG-GCN layers} 

To comprehensively evaluate the impact of the number of layers of MDG-GCNs on gait recognition performance, we systematically varied the GCN depth from 2 to 14 layers. As the MDG-GCN increases in depth, the model's ability to capture complex joint interactions improves, allowing it to better model the spatial and relational structures inherent in gait data. However, when the number of layers exceeds 12, the accuracy of the model reaches saturation and subsequently decreases, indicating potential overfitting. According to the experimental results shown in the Tab. \ref{table_gcn}, recognition accuracy increases with the number of layers up to 12, achieving an optimal performance of 94.34\%. Beyond 12 layers, the accuracy begins to decline.

\paragraph{Effectiveness of Non-linear Periodic Dynamic Learning}
To further assess the contribution of the non-linear periodic dynamic learning component, we conducted an ablation study by removing this module from our model. The recognition results are shown in Tab. \ref{table_1}, where w/o Temp. means the results without this module. Compared to the complete model, there is a significant performance drop (accuracy drops from 94.34\% to 89.94\%). This indicates that gait recognition relies heavily on analyzing the repetitive motion cycles of walking actions, such as the alternating movements of legs and arms.

% \begin{table}[]
% \centering
% \renewcommand{\arraystretch}{1.2}
% \setlength{\tabcolsep}{8pt}
% \caption{Performance comparison with and without temporal information on our dataset. The model with Temp, achieves better results}
% \label{table_temp}
% \begin{tabular}{c|ccc}
% \toprule
%     & Accuracy(\%)      & \multicolumn{1}{c}{F1-score} & \multicolumn{1}{c}{AUC} \\ \hline
% w/o Temp. & 89.94   & 0.8953                        & 0.9978                  \\
% w/ Temp.  & \textbf{94.34} & \textbf{0.9428}        & \textbf{0.9990}                 \\ 
% \bottomrule
% \end{tabular}
% \end{table}

% \begin{table}[ht]
% \centering
% \renewcommand{\arraystretch}{1.2}
% \setlength{\tabcolsep}{8pt}
% \caption{Performance comparison of GCN with different layers on the gait recognition task. The 12-layer GCN achieves the best performance}
% \label{tab:gcn_results}
% \begin{tabular}{c|ccc} 
% \toprule
% Layers & Accuracy (\%) & F1-score & AUC \\
% \hline
% 2  & 81.13 & 0.8082 & 0.9623 \\
% 4  & 85.53 & 0.8449 & 0.9597 \\
% 6  & 89.31 & 0.8900 & 0.9753 \\
% 8  & 90.57 & 0.8992 & 0.9801 \\
% 10 & 92.45 & 0.9203 & 0.9729 \\
% 12 & \textbf{94.34} & \textbf{0.9428} & \textbf{0.9990} \\
% 14 & 91.19 & 0.9051 & 0.9795 \\
% \bottomrule
% \end{tabular}
% \end{table}
\section{FUTURE WORK}

This paper combines the nonlinear periodic characteristics\cite{yue2024solitons} of human gait with multi-scale joint relationships to build an efficient recognition framework. In the future, we plan to collect gait data\cite{9796582} from more individuals in diverse environments to provide key support for training personalized exoskeleton control\cite{zhou2021lower}\cite{kalita2021development} models. Relying on high-precision gait recognition and the spatiotemporal representation of gait\cite{filipi2022gait}\cite{liu2020drone}presented in this paper, we will capture individuals' unique gait features in real time and achieve precise personalized exoskeleton control\cite{shushtari2021online}\cite{catalan2021modular} through adaptive algorithms, thereby offering safer and more efficient rehabilitation or assisted walking experiences for different users.

\section{CONCLUSIONS}

In this paper, we propose a novel skeleton-based approach for gait recognition on our constructed dataset. Our method effectively captures complementary gait periodic temporal dynamics and cross-scale spatial relational dependencies, significantly enhancing gait recognition accuracy. We conduct comprehensive experiments on our dataset, demonstrating that our approach surpasses \textit{state-of-the-art} methods, achieving 94.34\%, accuracy in 94.28\% in F1-score, and 99.90\% in AUC.

\bibliographystyle{ieeetr}
\bibliography{ref}

\end{document}